\documentclass{article}

\PassOptionsToPackage{numbers, compress}{natbib}

\usepackage[preprint]{neurips_2020}

\usepackage[utf8]{inputenc} 
\usepackage[T1]{fontenc}    
\usepackage{hyperref}       
\usepackage{url}            
\usepackage{booktabs}       
\usepackage{amsfonts}       
\usepackage{nicefrac}       
\usepackage{microtype}      
\usepackage{graphicx} 
\usepackage{multirow} 
\usepackage{amsmath}
\usepackage{xspace}

\hypersetup{
    colorlinks,
    linkcolor={red},
    citecolor={red},
    urlcolor={blue}
}

\usepackage{caption}
\usepackage{subcaption}
\usepackage{xcolor}
\usepackage{stmaryrd} 
\usepackage{wrapfig}

\definecolor{gray2}{gray}{0.3}

\newcommand\gray[1]{{\color{gray}{#1}}}
\newcommand\grayb[1]{{\color{gray2}{#1}}}

\newcommand{\IR}{\mathbb{R}}
\newcommand{\vect}{\mathbf}

\makeatletter
\DeclareRobustCommand\onedot{\futurelet\@let@token\@onedot}
\def\@onedot{\ifx\@let@token.\else.\null\fi\xspace}

\def\eg{\emph{e.g}\onedot} 
\def\ie{\emph{i.e}\onedot} 
 
 \def\vs{\emph{vs}\onedot}

\makeatother

\title{
Overcoming Statistical Shortcuts\\
for Open-ended Visual Counting
}

\author{
  \hspace{0cm} \vspace{0.2cm}Corentin Dancette $^{1}$\thanks{Equal contribution}, 
  Remi Cadene $^{1}$\footnotemark[1],  
  Xinlei Chen $^{2}$,  
  Matthieu Cord $^1$ \\
  $^1$ Sorbonne Universit\'e, CNRS, LIP6, 4 place Jussieu, 75005 Paris,
  \\ $^2$ Facebook AI Research,
  \texttt{\{corentin.dancette, remi.cadene, matthieu.cord\}@lip6.fr}, 
  \\ \texttt{xinleic@fb.com}
}

\begin{document}

\maketitle

\begin{abstract}  
Machine learning models tend to over-rely on statistical shortcuts.
These spurious correlations between parts of the input and the output labels does not hold in real-world settings.
We target this issue on the recent open-ended visual counting task which is well suited to study statistical shortcuts.
We aim to develop models that learn a proper mechanism of counting regardless of the output label.
First, we propose the Modifying Count Distribution (MCD) protocol, which penalizes models that over-rely on statistical shortcuts.
It is based on pairs of training and testing sets that do not follow the same count label distribution such as the odd-even sets.
Intuitively, models that have learned a proper mechanism of counting on odd numbers should perform well on even numbers.
Secondly, we introduce the Spatial Counting Network (SCN), which is dedicated to visual analysis and counting based on natural language questions.
Our model selects relevant image regions, scores them with fusion and self-attention mechanisms, and provides a final counting score.
We apply our protocol on the recent dataset, TallyQA, and show superior performances compared to state-of-the-art models.
We also demonstrate the ability of our model to select the correct instances to count in the image. Code and datasets are available: \href{https://github.com/cdancette/spatial-counting-network}{github.com/cdancette/spatial-counting-network}
\end{abstract}

\vspace{-0.5cm}
\section{Introduction}
\vspace{-0.2cm}

The recent advances in computer vision \citep{krizhevsky2012alexnet,he2016deep} and natural language processing \citep{mikolov2013efficient} allowed the research community to tackle challenging tasks that combine vision and language \citep{kiros2015skipthoughts, karpathy2015deep, lu2016vrd}.
One of these tasks is open-ended visual counting. 
Its goal is to count the number of instances in an image given a question formulated in natural language. It extends visual counting tasks, which are focused on one type of instance \citep{sindagi2018survey} or on a limited set of instances (\eg, 80 different objects \citep{chattopadhyay2017counting}). Solving it could pave the way towards the next generation of counting systems that possess interactive interfaces with applications in biology \citep{lempitsky2010learning}, medicine \citep{briggs2009quality}, wildlife monitoring \citep{onoro2016towards}, smart cities \citep{onoro2016towards, lempitsky2010learning} and more.

Open-ended counting was first introduced as a sub-task of Visual Question Answering (VQA) \citep{antol2015vqa, goyal2017vqa2, krishna2017vgenome, kafle2017tdiuc, johnson2017clevr} where the goal is to answer any type of questions about an image.
An important problem of VQA models is that they tend to \emph{memorize} statistical shortcuts \citep{geirhos2020shortcut} (also called spurious or superficial correlations \citep{agrawal2018vqacp}, unwanted priors or biases \citep{ramakrishnan2018overcoming, cadene2019rubi, selvaraju2019taking, wu2019self, jing2020overcoming}) between parts of the inputs and the output labels instead of learning proper \emph{mechanisms}.
They reach acceptable results on testing sets that follow a similar distribution as their training set but their performance degrades significantly otherwise \citep{agrawal2018vqacp}. This issue makes them impractical in real-world settings (see also \citep{stock2017imagenet, geirhos2018imagenet, barbu2019objectnet, alcorn2019strike, ilyas2019adversarial, goodfellow2014explaining} for pointing out this issue on object recognition tasks).
Open-ended counting models are greatly inspired and often compared to VQA models \citep{zhang2018counter, benyounes2017mutan}. While they are developed on specialized datasets for open-ended counting \citep{chattopadhyay2017counting, trott2018interpretable, acharya2019tallyqa}, their proximity with VQA models makes them potentially subject to similar problems of statistical shortcuts. 

\begin{figure}[t]
    \centering
    \includegraphics[width=\linewidth]{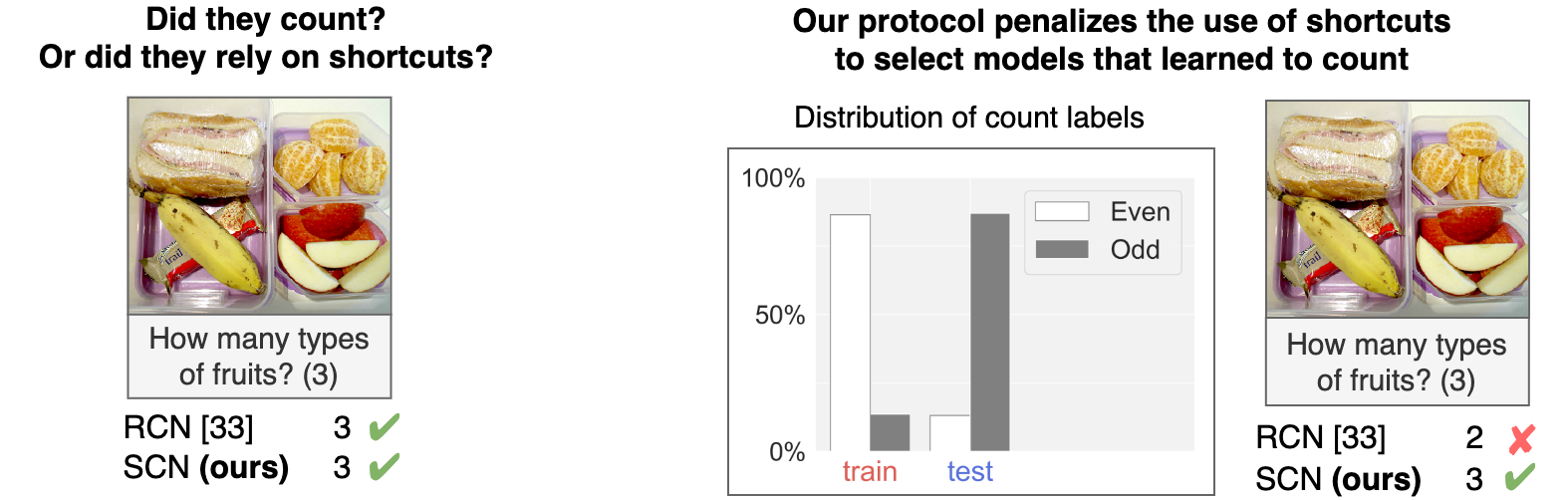}
    \caption{\label{fig:1} With the existing experimental protocols for open-ended counting, statistical shortcuts can be used to reach correct predictions without learning the underlying counting mechanisms.
    We propose a Modifying Count Distribution (MCD) protocol that penalizes models that over-rely on such shortcuts for counting.
    In this setting, the performances of state-of-the-art models (\eg RCN~\cite{acharya2019tallyqa}) are heavily impacted, while our proposed model, Spatial Counting Network (SCN) is more robust to distribution changes in counting.}
\end{figure}

In this paper, we first introduce a novel experimental protocol called Modifying Count Distribution (MCD). It is meant to select design choices that are useful for learning how to count instead of learning the shortcuts.
It is inspired by previous works on counting from cognitive science \citep{marcus1998rethinking, gross2009number} and on statistical shortcuts in VQA \citep{agrawal2018vqacp, teney2019actively}.
It consists in evaluating the ability to count of a given model on various training and testing sets that follow different count label distributions.
As shown in Figure~\ref{fig:1}, we evaluate the ability of a counting system trained on odd numbers to generalize on even numbers.
In this context, models must generalize to unseen or scarcely seen label counts and are heavily penalized for using shortcuts.
Inspired by \citep{trott2018interpretable}, we also evaluate their ability to correctly ground their final answer in the image. We use a standard object detection metric, and also introduce a new metric more suited to the counting task.

With this experimental protocol in mind, we introduce a novel model, Spatial Counting Network (SCN), dedicated to visual analysis and counting in the open-ended setting.
Contrarily to state-of-the-art approaches such as RCN \citep{acharya2019tallyqa} or Counter \citep{zhang2018counter} which are classification models, ours is a regression model.
This crucial design choice allows to learn more robust mechanisms by taking into account the structure of the output labels (ordered natural numbers), and by allowing the model to output counting values that have been scarcely or never seen in the training set.
Another important design choice is that our model assigns individual counting scores to image regions using fusion and self-attention mechanisms, before computing the final count number.
While ILRC \citep{trott2018interpretable} learns a hard selection of image regions using reinforcement learning, our model learns a soft selection in an end-to-end fashion. 
In addition, we introduce an entropy regularization term to enforce sparse regions scores. Our design choices guarantee a certain level of interpretability and help generalization on different count label distributions.

Our paper is designed along the following contributions. We first introduce the MCD protocol based on shifts in count label distribution between train and test sets. 
Secondly, we introduce an end-to-end learnable model for counting which integrates design choices allowing to learn robust counting mechanisms.
Finally, we apply our experimental protocol on the most recent and biggest open-ended counting dataset, TallyQA \citep{acharya2019tallyqa}. We pursue extensive experiments and show that our model performs better than current state-of-the-art models. We also validate our design choice by reporting improvements in grounding ability.

\section{Experimental protocol for open-ended counting}

An ideal experimental and evaluation protocol for open-ended counting should select models that learn the underlying \emph{mechanism} of counting, rather than models that rely on \emph{statistical shortcuts} \citep{geirhos2020shortcut}. These spurious correlations between parts of image-question inputs and the count labels allow models to perform well on pairs of training and testing sets that follow similar distributions but fail on real-world data due to a shift in distribution.

\subsection{Challenge of statistical shortcuts}

\paragraph{Detecting statistical shortcuts}
Real-world datasets often contain hidden statistical shortcuts that can be used to reach impressive performances. Detecting them is challenging.
A first approach consists in developing specific baselines that only rely on part of the inputs.
For instance, question-only models can be used to assess the existence of shortcuts between the question and the answer in VQA datasets \citep{antol2015vqa, goyal2017vqa2, agrawal2018vqacp}.
However, it is even more challenging to evaluate if state-of-the-art models over-rely on shortcuts.
Common approaches rely on expensive annotations \citep{das2017human} or on explainability methods \citep{stock2017imagenet, manjunatha2019explicit}.
Humans must then interpret if the displayed correlations are statistical shortcuts or not.

\paragraph{Penalizing statistical shortcuts}
Another approach consists in using testing splits that do not follow the training distribution to penalize models that learn these shortcuts instead of the proper mechanism. It simulates the kind of shifts in distribution that can potentially be encountered when deployed in real-world scenarios.
For instance, VQA-CP datasets \citep{agrawal2018vqacp} are built by re-organizing the training and testing sets of original VQA datasets, changing the distribution of answers per question type.
We propose a similar approach for open-ended counting datasets.
We introduce strategies to shift the label count distribution between the original training and testing sets. In this context, models must generalize to unseen or scarcely seen label counts and are heavily penalized for using shortcuts.
Finally, we select models according to their robustness in shifts in distribution. 

\subsection{Modifying Count Distribution protocol}

\begin{figure}[t]
    \centering
    \includegraphics[width=\linewidth]{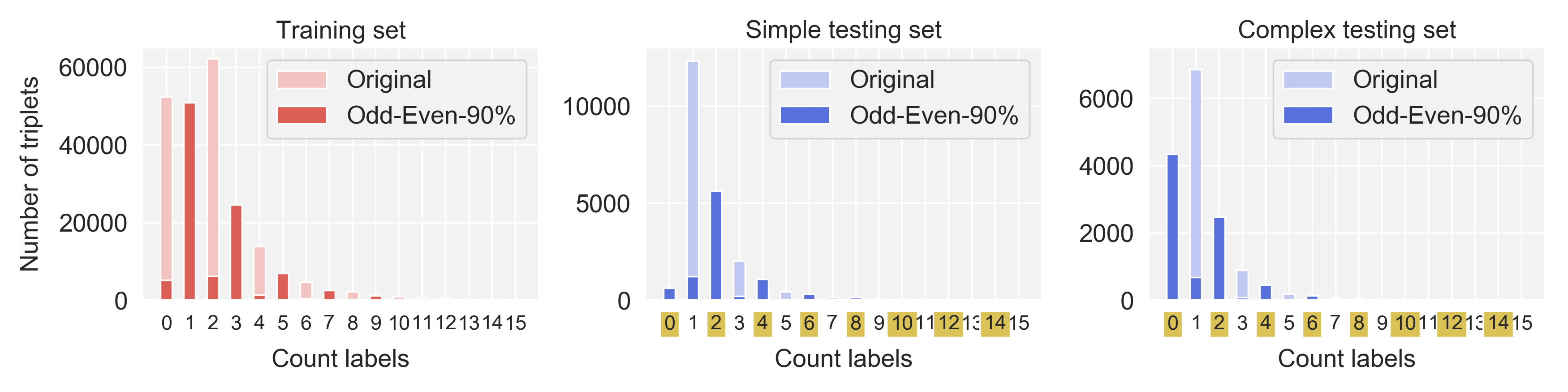}
    \caption{Number of triplets per count label on the TallyQA training set, testing set of simple questions and testing set of complex questions. Bar plots of the Odd-Even-90\% strategy  (in strong color) are displayed over the ones of the original TallyQA datasets (in light color). Models that over-rely on statistical shortcuts are penalized when evaluated on the even count labels (in yellow).}
    \label{fig:2}
\end{figure}

We now describe our experimental protocol, Modifying Count Distribution (MCD). It allows to penalize models that over-rely on statistical shortcuts without any need for external annotations or human supervision. Its goal is to select models that have learned a more robust counting mechanism.

\paragraph{Odd-Even-$p\%$ and Even-Odd-$p\%$ strategies}
Given a pair of training and testing sets made of \textit{image-question-label} triplets following similar distributions, we introduce strategies to produce a shift in distribution of count labels. 
The Odd-Even-$p\%$ generates unbalanced pairs by removing a percentage $p\%$ of triplets associated to an even label from the training set and removing the same percentage $p\%$ of triplets associated to an odd label from the testing set.
We control the amount of statistical shortcuts that can potentially be learned by varying $p\%$ from 0 to 100.
On the extreme sides, Odd-Even-$0\%$ generates the original pairs, while Odd-Even-$100\%$ generates a training set with no even count labels and a testing set with no odd count labels (i.e., a zero-shot setting).
Figure~\ref{fig:2} displays the shift in distribution obtained when applying the Odd-Even-$90\%$ strategy on the TallyQA training and testing sets.
The Odd-Even-$90\%$ is our strategy of choice because it introduces a large shift in distribution of count labels, while allowing classification models to learn from every possible answer.
Similarly, we introduce the Even-Odd-$p\%$ strategies to generate unbalanced training and testing sets which are mostly composed of triplets associated to even and odd count labels respectively.
As shown in the supplementary materials, all of our strategies produce a small shift in distribution of images and questions, which is important to only evaluate the impact of a shift in count labels.

\paragraph{Validation set}
As raised by \citep{teney2020value}, similar protocols \citep{agrawal2018vqacp} often select models based on their performance on the testing set only.
This bad practice encourages adaptive over-fitting \citep{dwork2015preserving} on the testing set distribution.
We address this common issue by introducing a validation set.
Given a pair of unbalanced training and testing sets, we build their associated validation set as a held-out subset of the training set. We use it to tune hyper-parameters and perform early-stopping to not reveal any information on the testing set distribution.

\section{Spatial Counting Network}
We now describe our model, Spatial Counting Network (SCN).
It contains inductive biases to encourage the learning of the counting mechanism, and avoid learning statistical shortcuts.
Our model uses multi-modal fusion and self-attention to assign counting scores to individual image regions, which allows the final accumulated count number to be spatially grounded.
In order to generalize to modified count distributions, we use a regression loss to train our model (as opposed to a classification loss \citep{zhang2018counter, acharya2019tallyqa}), and use entropy regularization to encourage the counting of natural numbers (as opposed to making discrete decisions trained with reinforcement learning \citep{trott2018interpretable}).

\begin{figure}[t]
    \centering
    \includegraphics[width=0.95\linewidth]{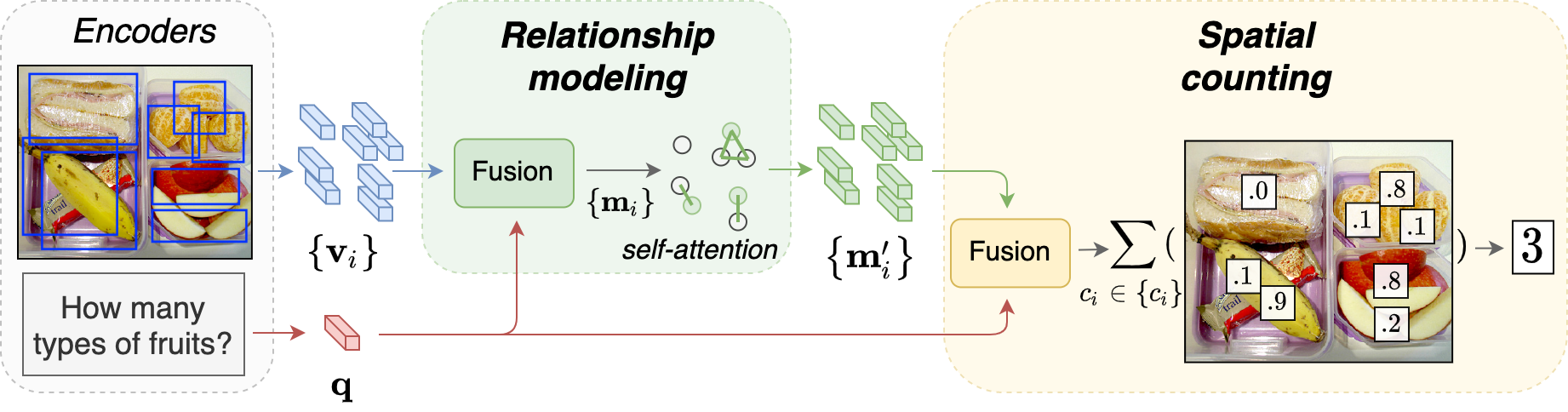}
    \caption{\label{fig:network} \textbf{Spatial Counting Network}. It takes an image and a counting question as inputs and outputs a count label. It is built upon object detectors to extract a set of spatially localized visual representations based on bounding boxes. Each of them is modified according to the question and their neighborhood until a counting score is obtained. The score indicates the presence (\eg ${\approx}1$) or absence (\eg ${\approx}0$) of a corresponding instance in the bounding box. The final count prediction is produced by simply summing up all the scores.}
\end{figure}

\paragraph{Overview} An overview of our model is shown in Figure~\ref{fig:network}. Formally, given a dataset $\mathcal{D}$ consisting of $n$ triplets $(v,q,c)$ with $v \in \mathcal{V}$ an image, $q \in \mathcal{Q}$ a natural language question and $c \in \mathbb{Z}^{+}$ a count label corresponding to the number of instances (non-negative) in the image, the goal is to learn a mapping $f_\theta{:}\mathcal{V}{\times}\mathcal{Q}{\rightarrow}\mathbb{Z}^{+}$ where $\theta$ denotes learnable parameters. Our model builds such a mapping by first encoding both inputs and fusing them, which we detail next.

\paragraph{Encoders and multi-modal fusion}
As shown in the first block of Figure~\ref{fig:network}, the model uses two encoders to produce vectorized representations for image $v$ and question $q$. For image $v$, a pre-trained object detector \citep{anderson2018bottom} is applied to transform the raw pixels to a set of $n_v$ spatially located vectors, with each vector $\vect{v}_i \in \IR^{d_v}$ encoding the semantic content of a region (or bounding box) within the image.
We project coordinates of each region into vectors of $d_v$ dimensions and sum them to their associated $\vect{v}_i$.
For $q$, we use skip-thought vectors \citep{kiros2015skipthoughts} to obtain its representation $\vect{q} \in \IR^{d_q}$.
We then merge each $\vect{v}_i$ with $\vect{q}$ using a multi-modal fusion module from \citep{kim2017mlb}, resulting in a new set of vectors $\{\vect{m}_i\}_{i\in\{1,\ldots,n_v\}}$ ready for relationship modeling and spatial counting, to be discussed below.

\paragraph{Relationships modeling} Since the set of bounding boxes used in encoding images can overlap, one core challenge for correct counting is to \emph{de-duplicate} boxes \citep{zhang2018counter, trott2018interpretable} that are assigned to the same instance. We address this by modeling general relationships among $\{\vect{m}_i\}$ using self-attention \citep{vaswani2017attention}, letting the model learn this mechanism. Specifically, a single-head attention module is applied on $\{\vect{m}_i\}$, yielding (for each region $i$) a contextualized representation $\vect{r}_i$, which is then element-wise summed with $\vect{m}_i$. The resulting vectors are denoted as $\{\vect{m}^{\prime}_i\}$. Beyond de-duplication, modeling pair-wise relationships could also be helpful for complex questions that require grouping regions (\eg `How many types of fruits ?') or spatial reasoning (\eg `How many cats are under the table ?').

\paragraph{Spatial counting} 
After relationship modeling, the resulting $\{\vect{m}^{\prime}_i\}$ vectors are then again fused \citep{kim2017mlb} with the question representation $\vect{q}$, and produce a counting score $c_i$ for each region via sigmoid activation. Finally, the global count output $\hat{c}{=}\sum_{i}{c_i}$ is a simple summation of all the individual counting scores. We name our model \textit{Spatial Counting Network}, because each and every count is explicitly grounded to a spatial region and allows for easy interpretation and visualization. 

While the above-described model encapsulates general components like multi-modal fusion and relationship modeling for open-ended counting, we would like to highlight two design choices that are important for improving its generalization to modified count distributions, described next.

\paragraph{Regression, not classification}
First, unlike many state-of-the-art counting models \citep{zhang2018counter, acharya2019tallyqa} (and general VQA models, including large-scale pretrained vision-and-language models \citep{lu2019vilbert,tan2019lxmert}) that treat count numbers as classification labels, we state they should be interpreted as actual numbers and directly train the model to regress the final output $\hat{c}$ to the ground truth count label $c$.
We choose the standard Mean Squared Error (MSE) as the loss: 
\begin{equation}
    \label{Eq:1}
    \mathcal{L}_{MSE}(\theta;\mathcal{D}) = \frac{1}{n} \sum_{(v,q,c) \in \mathcal{D}} (\hat{c} - c)^2.
\end{equation}
During testing, we round the fractional value $\hat{c}$ to its nearest integer to complete the mapping $f_\theta(v,q)$ to count labels. 
This loss is suited towards counting, as it takes advantage of the natural order of the count labels.
It also allows our model to output count labels that were not seen during training, which is beneficial when the testing set follows a different distribution of count labels.

\paragraph{Entropy regularization}
Second, although regression is a natural choice for number-related tasks, directly applying it to open-ended visual counting can be disadvantageous, because it attempts to model the entire output counting range (\ie $\hat{c}$ can be any real values between $0$ and $N$) and doesn't take advantage of the fact that \emph{all the count labels are integers}. One way to fix this is through reinforcement learning \citep{trott2018interpretable} which selects regions one by one, but the resulting objective function is hard to optimize directly. Here we propose an alternative solution by simply imposing a binary entropy regularization term per-region:
\begin{equation}
    \label{Eq:2}
    \mathcal{L}_{H}(\theta;\mathcal{D}) = -\frac{1}{n} \sum_{(v,q,c) \in \mathcal{D}} \left[\frac{1}{n_v} \sum_{i=1}^{n_v} c_i \log(c_i) + (1 - c_i) \log(1 - c_i)\right],
\end{equation}
which essentially encourages each sigmoid output $c_i$ to be close to $0$ or $1$. 
Intuitively, it means for each region, there is either one whole object, or none -- it won't be fractional (\eg 0.5). 
This regularization not only enforces the final count $\hat{c}$ to be close to integers (since $\hat{c}$ is produced by summing up scores that are close to $0$ or $1$), but also benefits \emph{grounding} the final count in the image (since it significantly reduces the chance of multiple overlapping regions being assigned some fractional value and summing up to be an integer count), which in turn helps generalization.

Combining MSE and entropy regularization, our final training loss is defined as: $\mathcal{L}{=}\mathcal{L}_{MSE} + \lambda \mathcal{L}_H$, where $\lambda$ is a fixed hyperparameter. We use $\lambda = 1$, which allows to reach the best performance in our context.

\section{Experiments}

\paragraph{Experimental setup}
We extensively use TallyQA \citep{acharya2019tallyqa}, the recent and biggest open-ended counting dataset. Its training set contains 130K real images from COCO \citep{lin2014microsoftcoco} and Visual Genome \citep{krishna2017vgenome}. Each image is associated with questions and count labels for a total of $\sim$250K triplets. 
It comes with a testing set of $\sim$23K simple questions and $\sim$16K complex questions.
Simple questions only require an object detection ability, while complex questions require abilities to detect relationships between objects, their attributes, spatial reasoning, and more \citep{acharya2019tallyqa}.
We compare our proposed model to state-of-the-art approaches and strong baselines following our experimental protocol which penalizes models for relying on statistical shortcuts.
Importantly, we do not incorporate knowledge about the testing set distribution such as sampling or weighting triplets based on their count labels.
We report only accuracy scores, as the RMSEs follow the same trend.
We then further study the impact of entropy regularization on the ability to select image regions that are important for counting. Implementation details are provided in the supplementary materials.

\subsection{State-of-the-art comparison using our MCD protocol}
\vspace{-0.3cm}
\begin{table}[!ht]
    \centering
    \caption{State-of-the-art comparison on a modified version of TallyQA using our Odd-Even-90\% strategy.
    90\% of the even labels and odd labels have been removed from the original training and testing sets respectively. 
    We report the final accuracy on the two testing sets: simple or complex. 
    We also report accuracy on the validation set which follows the training distribution and is used for early-stopping. Parameter counts are in millions.\label{tab:1}}
    \vspace{1mm}
    \begin{tabular}{l cc c c}
    \toprule
     & \multicolumn{2}{c}{Testing set}
     & \multirow{2}{*}{\gray{Validation set}}
     & \multirow{2}{*}{\# of Parameters}\\
      \cmidrule(lr){2-3} 
    & Simple & Complex & & \\
    \midrule
    Q-Only \citep{acharya2019tallyqa}   & 13.14  & 28.97 & \gray{53.78} & 28 M \\ 
    I-Only \citep{acharya2019tallyqa}  & 12.63 & 6.05 & \gray{53.77} & 4 M \\ 
    Q+I \citep{acharya2019tallyqa}  & 17.55 & 24.77 & \gray{61.08} & 30 M \\ 
    MUTAN \cite{benyounes2017mutan}  &  18.58 & 26.20 & \gray{64.33} & 58 M \\
    Counter  \citep{zhang2018counter}   & 17.99 & 20.43 & \gray{67.50} &  12 M  \\ 
    RCN \citep{acharya2019tallyqa}  & 30.64 & 26.69 & \textbf{\gray{69.53}} & 47 M \\ 
    \midrule
    RCN Regression  & 37.25 & 28.55 & \gray{61.41}& 47 M\\ 
    SCN (ours)  & \textbf{48.22} & \textbf{32.54} & \gray{54.78} & 52 M\\
    \bottomrule
    \end{tabular}
    \vspace{-0.3cm}
\end{table}

\paragraph{Main results}
In Table~\ref{tab:1}, we compare our model against state-of-the-art approaches and other reported models on TallyQA.  Notably, RCN \citep{acharya2019tallyqa} and Counter \citep{zhang2018counter} are specifically designed to answer counting questions. Scores for SCN are averaged over three runs, with a variance for simple and complex of 0.3 and 1.1 respectively.
We train and evaluate each model on a modified version of TallyQA using our Odd-Even-90\% strategy.
Models that over-rely on statistical shortcuts are expected to perform well on its validation set since it follows the training set distribution but suffer from a large loss in accuracy on the testing sets.
Interestingly, RCN and Counter reach a high accuracy of 69.53\% and 67.50\% on the validation set, but suffer from huge losses of -38.89 and -49.51 accuracy points respectively on the simple testing set. We observe similar losses on the complex testing set.
On the contrary, our model reaches the best accuracy of 48.22\% on simple questions and 32.54\% on complex questions, with gains in accuracy of +17.58 and +5.85 respectively over RCN \citep{acharya2019tallyqa} (third last row) which is the state-of-the-art model and has a similar number of parameters (52M \vs 47M).

\paragraph{Impact of regression loss}
A notable difference between our model and state-of-the-art models such as RCN and Counter is that they are trained using classification instead of regression. For fair comparisons, 
we isolate the contribution of this design choice by introducing RCN Regression, which is a modified RCN that outputs a real number before rounding and is trained using the MSE loss.
In Table~\ref{tab:1} (second last row), we report an accuracy of 37.25\% and 28.55\% on simple and complex questions from the testing set with gains of +6.61 and +1.86 accuracy points respectively over the original version of RCN.
Compared to RCN, we note a smaller loss in accuracy between the validation and testing sets with -24.16 accuracy points on the simple questions against -38.89.
These good performances indicate that regression models are a better design choice to avoid learning statistical shortcuts.
However, other design choices allow our model to reach further gains with +10.97 and +3.99 accuracy points on simple and complex questions against RCN Regression. These gains are significantly higher than those resulting in the introduction of the regression alone (+6.61 and +1.86).

\subsection{Detailed comparison using our MCD protocol}

\begin{figure}[h!]
    \centering
    \includegraphics[width=\linewidth]{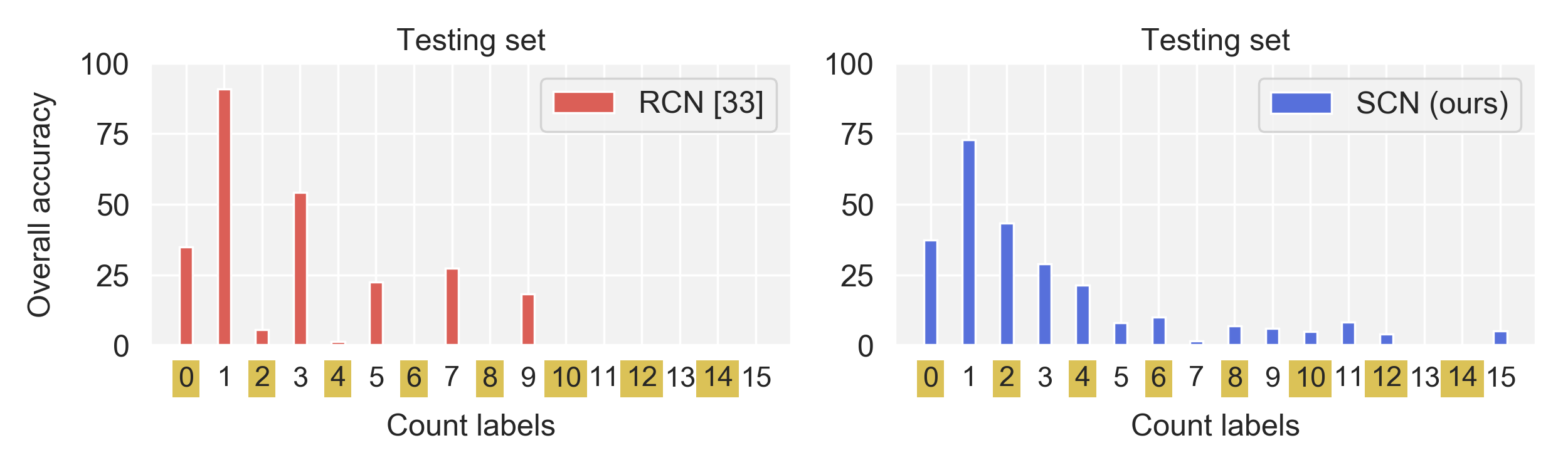}
    \vspace{-0.7cm}
    \caption{Comparison between our model and RCN \cite{acharya2019tallyqa} on our modified version of TallyQA using our Odd-Even-90\% strategy.
    Our model reaches significantly better overall accuracy on even labels (in yellow). These count labels are meant to penalize models that over-rely on statistical shortcuts.}
    \label{fig:4}\vspace{-0.3cm}
\end{figure}

\paragraph{Difference in accuracy per count label}
Gains in accuracy could be due to different patterns such as an important gain on only one count label or small gains on all of them. We study this in Figure~\ref{fig:4}, where we display a fine-grained comparison between our model and RCN according to their overall accuracy per count label.
Interestingly, we report a higher accuracy on even count labels which are less represented in the training set and a lower accuracy on odd count labels which are more represented in the training set. We also report much smaller differences in accuracy between adjacent count labels, compared with RCN. For instance, we report a loss of -29.56 accuracy points between label 1 and 2 compared to -85.15 with RCN.
Overall, there is much less variation in our model between even and odd count labels.
These results suggest that our design choices are useful to learn a proper mechanism of counting which helps to generalize to a different distribution of count labels.

\begin{figure}[!h]
    \centering
    \includegraphics[width=0.94\linewidth]{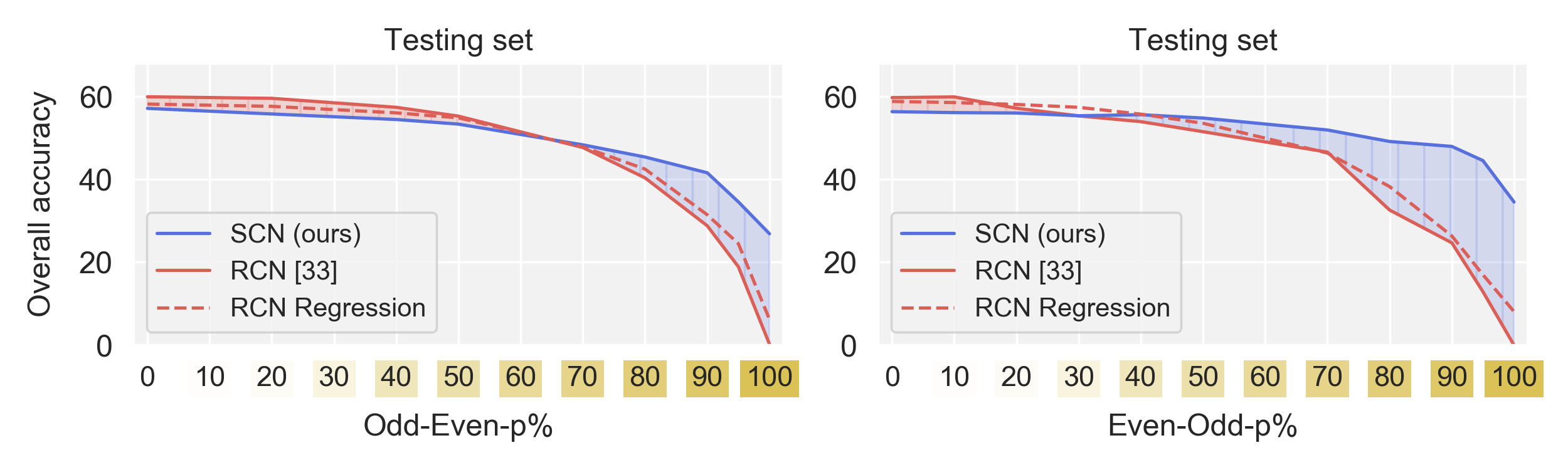}
    \vspace{-0.3cm}
    \caption{Comparison between our model, RCN \cite{acharya2019tallyqa} and its regression variant on various versions of TallyQA using our Odd-Even-p\% and Even-Odd-p\% strategies.
    p\% controls the shift in distributions between the training and testing sets (with the original distribution when p = 0).
    Models that over-rely on statistical shortcuts (\eg original RCN) are strongly penalized when p\% is high (yellow gradient).}
    \label{fig:5}\vspace{-0.3cm}
\end{figure}

\paragraph{Difference in accuracy on various shifts in distribution}
In Figure~\ref{fig:5}, we compare our model against the state-of-the-art model RCN for open-ended counting, and its regression version, according to their overall accuracy on a variety of datasets that can be generated with our Odd-Even-p\% and Even-Odd-p\% strategies. We vary p from 0 to 100 to go from no shift in distribution to the highest shift.
We show that while RCN reaches a slightly better accuracy when the shift in distribution is moderate (\eg, p < 60), our model reaches significant and consistent gains when the shift is bigger (\eg p > 60).
As expected, we report larger gains over RCN ranging from +12.78 accuracy points to +34.52 on datasets that possess the most important shift in distributions (e.g. p > 80). We see similar gains over RCN Regression.

\subsection{Study of the grounding ability}

\paragraph{Comparison on COCO-Grounding}
We measure the grounding ability as a proxy to evaluate if models have learned the proper counting mechanism, and to assess the interpretability of our model.
To this end, we specifically design a dataset named \emph{COCO-Grounding}, and a grounding metric for open-ended counting called \emph{GroundP}. Both are detailed in the supplementary material and are publicly available. Unlike \citep{trott2018interpretable}, GroundP can be usable in future work where models may use different visual features than ours. Intuitively, the GroundP metric measures the proportion of the total score that is correctly located in the ground truth bounding boxes. 
In Table~\ref{tab:tallyqa_grounding}, we first compare our best model against the state-of-the-art on our GroundP metric and on the mean average precision $\mathrm{AP}$, which is a standard object detection metric. We use a $\mathrm{AP}$ with a detection threshold of 0.2.
Both models have been trained on the original TallyQA dataset. As expected, we report a lower accuracy since models that over-rely on statistical shortcuts are not penalized on this dataset.
We report the best performances on grounding, and a gain of +24.3 and +9.1 points respectively over our retrained version of Counter.
We do not compare against RCN \citep{acharya2019tallyqa}, because it does not internally associate counting numbers to regions of the image.
We also report a gain of +8.7 and +4.7 points respectively over our model optimized without entropy regularization. These results justify the effectiveness of our regularization (Eq.~\ref{Eq:2}) for counting.

\begin{table}[!ht]
    \centering
    \caption{\label{tab:tallyqa_grounding}Grounding ability of models trained on original TallyQA. We report standard $\mathrm{AP}$ and our GroundP metric on COCO-Grounding dataset (see supplementary materials for details). Our retrained Counter*~\citep{zhang2018counter} reaches a different balance between simple and complex sets.}
    \vspace{1mm}
    \begin{tabular}{l cc cc cc}
    \toprule
    & \multicolumn{2}{c}{COCO-Grounding}
    & \multicolumn{2}{c}{\grayb{TallyQA Simple}}
    & \multicolumn{2}{c}{\grayb{TallyQA Complex}} 
     \\
    \cmidrule(lr){2-3}  \cmidrule(lr){4-5} \cmidrule(lr){6-7}
    & GroundP & $\mathrm{AP}$ & \grayb{Acc.} & \grayb{RMSE} & \grayb{Acc.} & \grayb{RMSE}  \\
    \midrule
    SCN (ours)  & \textbf{51.4} & \textbf{22.7} & \grayb{63.23} &  \grayb{1.09} & \grayb{47.01} & \grayb{1.46} \\
    SCN w/o entropy & 42.7 & 18.0 & \grayb{63.88} & \grayb{1.07} & \grayb{47.03} & \grayb{1.45}   \\
    Counter* \citep{zhang2018counter} & 27.1 & 13.5 &  \grayb{65.39} & \grayb{1.27} & \grayb{53.49}  & \grayb{1.51}  \\
    \midrule
    RCN \citep{acharya2019tallyqa} & - & - &  \grayb{71.8} & \grayb{1.13} & \grayb{56.2}  & \grayb{1.43}  \\
    Counter \citep{zhang2018counter} & - & -  & \grayb{70.5} & \grayb{1.15} & \grayb{50.9} & \grayb{1.58} \\
    \bottomrule \\
    \end{tabular}
    \vspace{-0.5cm}
\end{table}

\paragraph{Qualitative study}

\begin{figure}[!h]
    \centering
    \includegraphics[width=0.90\linewidth]{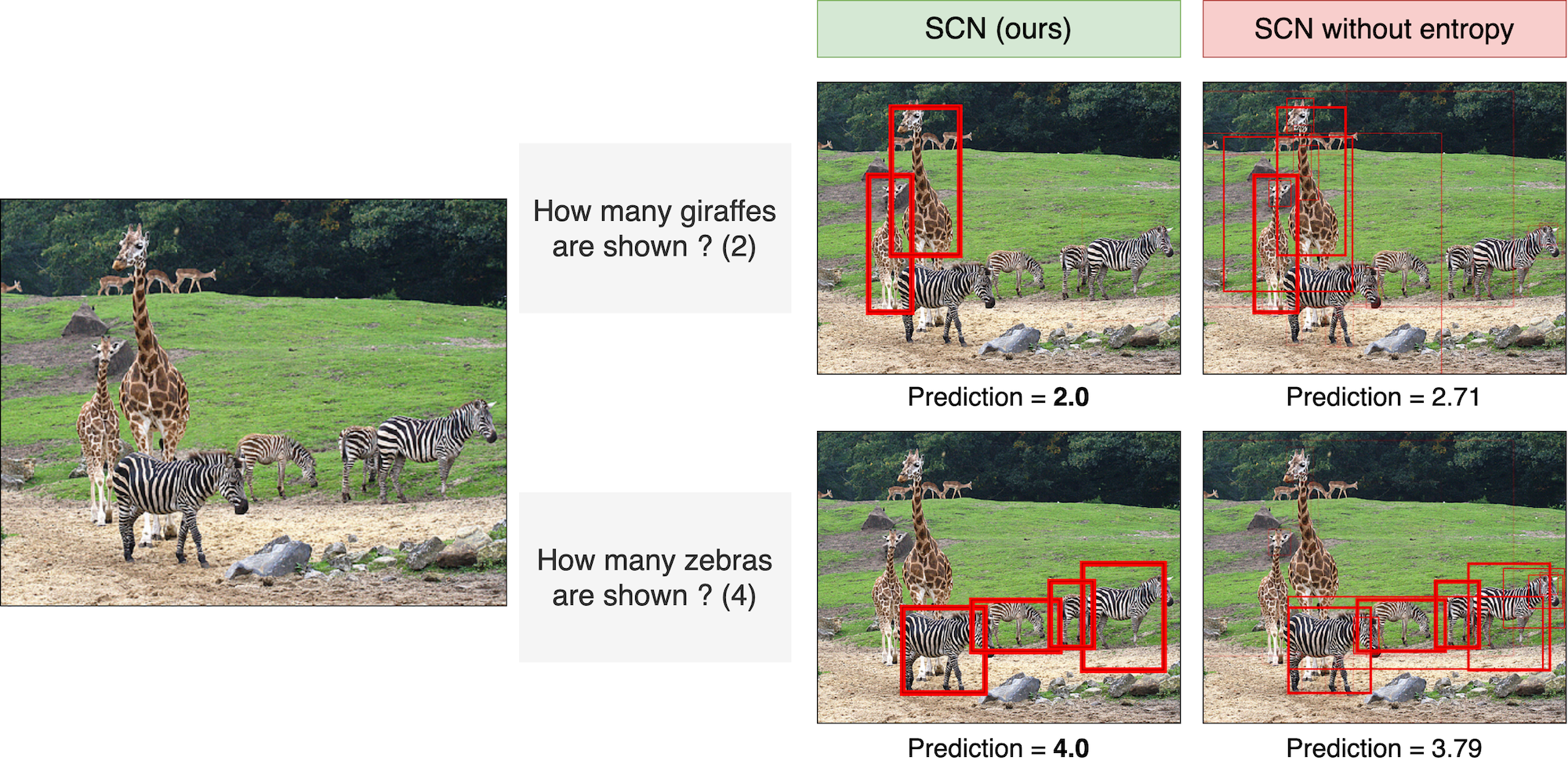}
    \vspace{-0.1cm}
    \caption{Qualitative comparison between our model with and without entropy regularization regarding their ability to select the correct regions to count and to provide correct predictions. Red bounding boxes are shown with bolded borders when their associated $c_i$ is close to 1.}
    \label{fig:qualitative}
    \vspace{-0.1cm}
\end{figure}

In Figure~\ref{fig:qualitative}, we display representative examples of outputs of our model with (on the left) and without (on the right) entropy regularization. Both versions are the same to those compared in Table~\ref{tab:tallyqa_grounding}.
We first compare both models on their ability to get the correct prediction for the question `How many giraffes are shown?'.
We display bolded red bounding boxes around objects when their associated count value $c_i$ is closed to 1.
We find our model trained with entropy selects the correct two regions of giraffes.
On the other hand, our model without entropy fails to distinguish duplicates and associates fractional values to multiple regions that possess giraffes. It also predicts the wrong count label (2.71 is rounded to 3).
We report similar observations for the question `How many zebras are shown?'. Our entropy regularization strategy is thus critical to improve the interpretability of our model.
More examples can be found in supplementary materials.

\vspace{-0.2cm}
\section{Conclusion}

We propose an experimental protocol, called Modifying Count Distribution (MCD), to penalize open-ended counting models that over-rely on statistical shortcuts. It generates various modified dataset versions where the distributions of even and odd count labels are different between the training and testing sets, while keeping similar distributions of words and images.
We then introduce a model, Spatial Counting Network (SCN), which encompasses important design choices that help to overcome statistical shortcuts. Specifically, it models region relationships and associates a score to each region before summing them to the final predicted count.
It is trained with a regression loss and a regularization that minimizes the binary entropy of each score.
We evaluate SCN against state-of-the-art models and report more robustness to distribution changes.
We also show that our entropy-based regularization strategy has a beneficial impact on grounding ability.
For future work, we plan to extend our experimental protocol to more general machine learning problems.

\newpage

\paragraph{Broader impact}
We develop a framework that aims at reducing the undesired learning of statistical shortcuts, or unwanted biases, from the training data. This is a common and important issue in vision-and-language tasks, and more generally in machine learning. Reducing the learning of shortcuts is essential if we aim to use those models in the real world, where the data distribution does not necessarily follow the training data distribution. 
It is also related to the algorithm fairness, \ie the development of models that are independent of some sensitive variables. We believe our approach could be used in similar settings where shortcuts can harm fairness.
Finally, our work is a step towards better interpretability for counting models. Interpretability is an important characteristic of models that can have a positive impact on trust towards those systems.

\paragraph{Acknowledgment}

We would like to thank Manoj Acharya and Hisham Cholakkal for their availability and kindness when answering our questions.

The effort from Sorbonne University was partly supported within the Labex SMART supported by French state funds managed by the ANR within the Investissements d’Avenir programme under reference ANR-11-LABX-65, and partly funded by grant DeepVision (ANR-15-CE23-0029-02, STPGP-479356-15), a joint French/Canadian call by ANR \& NSERC. This work benefited from the Jean-Zay cluster.


\newpage
\section{Supplementary materials}

We design the supplementary materials along the main paper sections.
In section \ref{sec:mcb-protocol}, we provide details about the training, validation and testing sets generated with our Odd-Even-p\% and Even-Odd-p\% strategies on the TallyQA dataset and show that they lead to small shifts in distribution of words and visual concepts while allowing big shifts in distribution of count labels.
In Section \ref{sec:implem-detail}, we provide implementation details about models used in this study to ease the reproducibility of our results.
In Section \ref{sec:grounding-details}, we provide details about the grounding metrics.
In Section \ref{sec:additional-results}, we provide additional experiment results. We show that our SCN model reaches higher performances than standard baselines. We also display qualitative results.
In Section \ref{sec:ml-reproducibility}, we directly address questions on reproducibility from the NeurIPS 2020 community. We also join the code in a zip file. Details about the code can be found in the README.md file inside the zip.

\subsection{Details about the MCB protocol}
\label{sec:mcb-protocol}

\paragraph{Training, validation and testing sets statistics}

Before applying our ablation strategies (Odd-Even-p\% and Even-Odd-p\%) on the TallyQA dataset, we first build a validation set by removing 10\% of images from the original training set.
All \textit{image-question-count} triplets that possess those images are set aside to build the validation set.
We then apply a chosen ablation strategy to each set so that the training and validation sets follow the same count label distributions while the testing set follow a different one.
For instance, the Odd-Even-90\% strategy removes 90\% of triplets associated to even count labels from the training set, 90\% of triplets associated to even count labels from the validation set, and 90\% of triplets associated to odd count labels from the testing set.
In Table~\ref{tab:num-examples-tallyqa-oe}, we display the number of odd and even triplets in each set when we apply the Odd-Even-p\% strategy on TallyQA with various value of $p$.
In Table~\ref{tab:num-examples-tallyqa-eo}, we display the number of triplets in each set when we apply the Even-Odd-p\% strategy on TallyQA with various value of $p$.

\begin{table}[!h]
    \centering
    \begin{tabular}{ccccccc}
    \toprule
    &  \multicolumn{2}{c}{Training set} & \multicolumn{2}{c}{Validation set} 
    & \multicolumn{2}{c}{Testing set} \\
      \cmidrule(lr){2-3}  \cmidrule(lr){4-5}  \cmidrule(lr){6-7}  
    p\% & Odd & Even & Odd & Even & Odd & Even \\ 
    \midrule
    0 \%   & 87,289 & 137,102 & 9,635 & 15,292 & 23,138 & 15,451 \\
    50 \%  & 87,289 & 68,549  & 9,635 & 7,644  & 11,565 & 15,451 \\
    90 \%  & 87,289 & 13,707  & 9,635 & 1,525  & 2,328  & 15,451 \\
    100\%  & 87,289 & 0       & 9,635 & 0      & 0      & 15,451 \\
    \bottomrule \\
    \end{tabular}
    \caption{Number of \textit{image-question-count} triplets  for each set generated by our Odd-Even-p\% strategy when applied on the TallyQA dataset (Odd-Even-0\% leads to the the original TallyQA distribution). Numbers of triplets for intermediate values of $p$ can be obtained with a linear interpolation.}
    \label{tab:num-examples-tallyqa-oe}
\end{table}

\begin{table}[!h]
    \centering
    \begin{tabular}{ccccccc}
    \toprule
    &  \multicolumn{2}{c}{Training set} & \multicolumn{2}{c}{Validation set} 
    & \multicolumn{2}{c}{Testing set} \\
      \cmidrule(lr){2-3}  \cmidrule(lr){4-5}  \cmidrule(lr){6-7}  
    p\% & Odd & Even & Odd & Even & Odd & Even \\ 
    \midrule
        0 \% & 87,289 & 137,102 & 9,635 & 15,292 & 23,138 & 15,451 \\
    50 \%    & 43,643 & 137,102 & 4,815 & 15,292 & 23,138 & 7,719 \\
    90 \%    & 8,725  & 137,102 & 969  & 15,292 & 23,138 & 1,551 \\
    100 \%   & 0      & 137,102 & 0     & 15,292 & 23,138 & 0 \\
    \bottomrule \\
    \end{tabular}
    \caption{Number of \textit{image-question-count} triplets  for each set generated by our Even-Odd-p\% strategy when applied on the TallyQA dataset (Even-Odd-0\% leads to the the original TallyQA distribution). Numbers of triplets for intermediate values of $p$ can be obtained with a linear interpolation.}
    \label{tab:num-examples-tallyqa-eo}
\end{table}

\paragraph{Shift in distribution of questions and visual concepts}

We compute the distributions of words from the questions and visual concepts in the images in various Odd-Even-p\% training sets, and compare them to the original distributions of TallyQA. To compute the words distribution, we proceed as follow. We first remove the common words  \textit{how, many, can, you, scene, picture, pictured, image, photo, there, are, seen, see, visible, shown, this, in, the, on, be, of, a, to} to only keep those that are associated to specific concepts in the images. We then compare the distributions using the Bhattacharyya coefficient \cite{bhattacharyya1946measure} -- a similarity metric which reaches 0 when there is no overlap between distributions, and 1 when both are the same. 
Similarly, we compute visual concepts distributions by using the categories assigned to every bounding box extracted from our pre-trained object detector \citep{anderson2018bottom}, and compare the distributions using the Bhattacharyya coefficient.
In Table~\ref{tab:word-distribution-distance}, we see that all coefficients are very close to 1, even for the datasets generated with the 100\% strategies, which confirms that our protocol l very little the distributions of words and visual concepts.

\begin{table}[!h]
    \centering
    \begin{tabular}{r cc}
    \toprule
        p\% & Words Similarity & Visual similarity \\
        \midrule
        0 \% & 1.0 & 1.0 \\
        50 \%  & 0.997 & 0.9999 \\
        90 \%  & 0.986 & 0.9996 \\
        100 \%   & 0.976 & 0.9995 \\
        \bottomrule \\
    \end{tabular}
    \caption{Bhattacharyya coefficients \cite{bhattacharyya1946measure}. Words and visual concepts similarity between each of our generated training sets using our Odd-Even-p\% strategy and the original TallyQA training set.}
    \label{tab:word-distribution-distance}
\end{table}

\subsection{Implementation details}
\label{sec:implem-detail}

We will release the Pytorch \cite{paszke2019pytorch} code to generate the datasets and reproduce our results. We will also release pre-trained models and configuration files with the following hyperparameters.

\paragraph{Our SCN model}
We use the common Faster R-CNN \cite{ren2015faster} pre-trained by \citep{anderson2018bottom} to extract object features from the image, and the common GRU language model pretrained by \citep{kiros2015skipthoughts} to extract language features from the question. To keep a similar number of parameters with the state-of-the-art RCN model \citep{acharya2019tallyqa}, We use hidden dimensions of 1500 for the multimodal embeddings $m_i$, 500 for the self-attention, 768 for both bilinear fusions, and use only one self-attention head.
We train our model for 30 epochs with the Adam optimizer \citep{kingma2015adam} and a learning rate of 2.e-5 which is decayed by 0.25 every 2 epochs, starting at epoch 15. The learning rate scheduling was tuned on the validation accuracy of the Odd-Even-90\% set. Importantly, for all other experiments, we use the exact same hyper-parameters.
We early stop training based on the highest accuracy computed on the validation set.
During training, we fix the $\lambda$ weight controlling the influence of the entropy loss to 1 in order to keep a similar order of magnitude between the gradients norm computed using the entropy loss and the gradients norm computed using the MSE loss. 
After obtaining our main results, we experimented values around 1 to assess its robustness. We report small variations in accuracy scores. 

\paragraph{RCN and RCN regression}
We follow the implementation and hyperparameters described in \cite{acharya2019tallyqa}. We create RCN regression by changing the output dimension of the last linear layer from 15 to 1. This allows us to train the model with a MSE regression loss instead of a classification loss. We use the same hyperparameters as RCN.

\subsection{Details about grounding experiments}
\label{sec:grounding-details}

\paragraph{COCO-Grounding dataset}
Similarly to the work done in IRLC \cite{trott2018interpretable}, we use the grounding ability as a proxy to evaluate the proper counting mechanism, and to assess the interpretability of models. To this end, we design COCO-Grounding, a dataset specifically designed to be usable in future work where visual features may be different than ours.
Our dataset is composed of the 4459 images from MSCOCO \citep{lin2014microsoftcoco} that can not be found in Visual Genome \citep{krishna2017vgenome} and importantly not in the TallyQA training set. Each MSCOCO image is annotated with bounding boxes around objects associated with a category among 80 classes of objects. We use these classes to automatically generate simple questions about a given image using the "How many \{class\}?" pattern. The answer to a question is a count label obtained by counting the number of bounding boxes associated to the given \{class\}.
We also generate questions associated to the count label 0 by sampling a random class among 80 that is not present on the image. 
We generate an equal number of 734 \textit{image-question-count} triplets associated to the count label 0, 1 and 2, and generate all possible triplets for higher count labels (with a maximum label of 15) to reach a total number of 3311 triplets over 2139 images. This subset will be publicly released.

\paragraph{Evaluation metrics}
Similarly to object detection models, our model can output bounding box predictions. We thus use a standard metric in object detection tasks \cite{everingham2015pascal, lin2014microsoftcoco} called mean average precision. It allows us to evaluate the ability of our model to detect the correct instances of objects to count in the image.
We also introduce a novel grounding metric specifically designed for open-ended counting. We refer to it as \texttt{GroundP} for Grounding Precision. It is derived from the grounding metric in \cite{trott2018interpretable}, with the difference that we use the ground truth bounding boxes as references to compute the grounding, instead of the bounding boxes extracted from the object detection model. This enables us to more accurately evaluate the grounding, and to compare models with different object detection models.
The metric consists in weighting the scores assigned by our model to each proposed bounding box by the portion of their size that overlaps with the ground truth bounding boxes.
More details are given in the next paragraph.

\paragraph{Details about our GroundP metric}
For each input triplet $m$, we have a set of ground truth bounding boxes $\mathrm{gt}^m_i$. We note $\mathrm{GT}^m$ the union of all those bounding boxes. It represents the total area of counted objects.
For this triplet $m$, our object detection model returns us $R$ region proposals $(b^m_i)_{i \in [1, R]}$. Our model returns, for each region, a score  $s^m_i$.  The final score is $C^m = \sum_i s^m_i$. For every proposed bounding box $b^m_i$, we compute its precision (the intersection with ground truth bounding boxes over its own area).
A precision of 1 means that the bounding box is totally in the ground truth area, whereas a precision of 0 means that is is totally disjoint.
\begin{align}
    p^m_i = \frac{\mathrm{Area}(\mathrm{GT}^m \cap b^m_i)}{\mathrm{Area}(b^m_i)}
\end{align}
We then weight the precision by the score $s^m_i$ assigned by our model to this region and sum over the regions to obtain our final score: 
$S^m = \sum_i s^m_i \cdot  p^m_i$.
We recall that our model's prediction is $C^m = \sum_i s^m_i$.
The final metric is computed by summing the results over all the images, and normalizing by the sum of predicted scores.
\begin{align}
S = \frac{\sum_{m \in D} S^m}{\sum_{m \in D} C^m}
\end{align}
This interpretation of the metric is straightforward: It represents the proportion of the final score that is correctly grounded in the image. A value of one would mean that all chosen objects (with a nonzero score) are in the ground truth bounding boxes.

\subsection{Additional results}
\label{sec:additional-results}

\paragraph{Baselines}

In Table~\ref{tab:supplementary-results}, we compare additional baselines against our SCN model on datasets generated with the Odd-Even-90\% strategy.
The \textit{Random (train)} baseline consists in randomly sampling the predicted count labels following their distribution in the training set. \textit{Random (test)} samples count labels following their distribution in the testing set. While this baseline leverages knowledge about the testing set distribution, SCN reaches significantly better accuracy scores with +16.88 accuracy points in simple questions and +4.76 against it.
The RCN + Uni. Sample. baseline is trained with a different triplets sampling strategy than RCN based on a uniform sampling of the count labels. By doing so, we seek to train a more robust RCN model against shift in distributions between training and testing sets. As expected, we report lower accuracy scores than RCN on the validation set with a loss of -4.09 point. Interestingly, we report lower scores on the simple testing set with a loss of -4.14 points and slightly better score on the complex testing set with a gain of +1.09. Overall, we report significantly lower scores than our SCN with -21.72 accuracy point on simple. These results show that the uniform sampling strategy is not suited to learn robust counting mechanisms.

\begin{table}[!h]
    \centering
    \begin{tabular}{l ccc}
    \toprule
      &  \multicolumn{2}{c}{Testing set}
     & \multirow{2}{*}{Validation set} \\
      \cmidrule(lr){2-3} 
         &  Simple & Complex  \\
    \midrule
       Random (train distribution)  & 11.04 & 9.21 & \gray{32.15} \\
       Random (test distribution) & 31.34 & 29.95 & \gray{10.37} \\
       RCN* \citep{acharya2019tallyqa} +  Uni. Sampl. & 26.50  & 27.78 & \gray{65.44} \\
       \midrule
        RCN* \citep{acharya2019tallyqa} & 30.64 & 26.69 & \gray{\textbf{69.53}} \\
        SCN (ours)  & \textbf{48.22} & \textbf{32.54} & \gray{54.78} \\
    \bottomrule \\
    \end{tabular}
    \caption{Comparison against additional baselines on a modified version of TallyQA using our Odd-Even-90\% strategy.
    90\% of the even labels and odd labels have been removed from the original training and testing sets respectively. 
    We report the final accuracy on the two testing sets: simple or complex.
    We also report accuracy on the validation set which follows the training distribution and is used for early-stopping.}
    \label{tab:supplementary-results}
\end{table}

\paragraph{Qualitative results}
\label{sec:qualitative-results}

\begin{figure}[h!]
    \centering
    \includegraphics[width=1.0\linewidth]{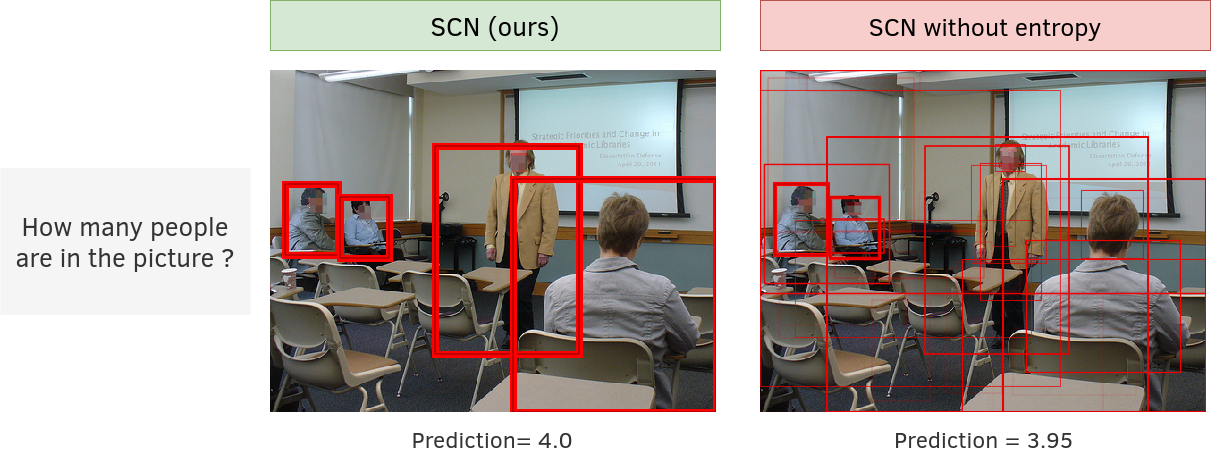}
    \caption{Qualitative comparison between our SCN model with and without entropy regularization regarding their ability to select the correct regions to count and to provide correct predictions. Red bounding boxes are shown with bolded borders when their associated $c_i$ is close to 1.}
    \label{fig:qualitative-people}
\end{figure}

\begin{figure}[th!]
    \centering
    \includegraphics[width=0.8\linewidth]{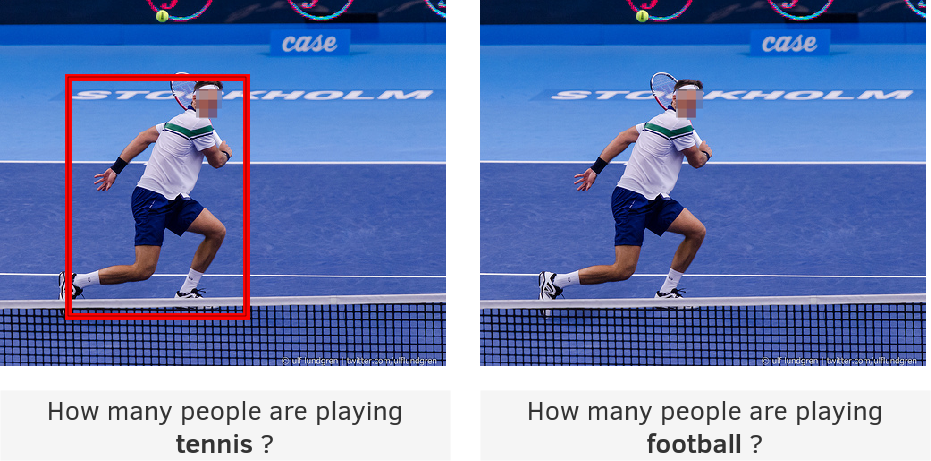}
    \caption{Counting scores produced by our SCN model for two complex questions on the same image. Red bounding boxes are shown with bolded borders when their associated $c_i$ is close to 1.}
    \label{fig:qualitative-tennis}
\end{figure}

In Figure~\ref{fig:qualitative-people}, we show the influence of our entropy loss. We compare our SCN model and a SCN model trained without entropy loss.
For the question 'How many people are in the picture?', we see that our SCN model selects the correct four regions, while our model trained without entropy fails to distinguish duplicates and associates fractional values to multiple regions that possess people.
I also leads to a fractional prediction of the count label.
In Figure~\ref{fig:qualitative-tennis}, we display two complex questions on the same image, and show that our SCN model is able to select an object (person) and filter according to an attribute (sport).


\vspace{2cm}
\subsection{ML reproducibility}
\label{sec:ml-reproducibility}

\textbf{The range of hyper parameters considered, method to select hyper-parameter configuration and specification of all hyper-parametrs used to generate results.}\\
See Section~\ref{sec:implem-detail}.

\textbf{The exact number of training and evaluation runs.}\\In Table~\ref{tab:1}, we report the mean over 3 runs with different seeds for our main result. We report the variation in the associated paragraph.
All other experiments are launched once.

\textbf{A clear definition of the specific measure or statistics used to report results.}\\We report standard Accuracy and RMSE which are widely used metrics, grounding metrics which are detailed in Section~\ref{sec:grounding-details}.

\textbf{A description of results with a central tendancy (eg mean) \& variation (eg error bars).}\\In Table~\ref{tab:1}, we report the mean over 3 runs with different seeds for our main result. We report the variation in the associated paragraph.

\textbf{The average runtime for each result, or estimated energy cost.}\\Our SCN model takes 10 hours to train on the original TallyQA and 5 hours on the dataset generated by the Odd-Even-100\%. It is the smallest dataset with about half the size of the original TallyQA.

\textbf{A description of the computing infrastructure used.}\\We train our model on a single GPU. We have access to several 12GB Titan X Pascal and 32GB Tesla V100.

\end{document}